\documentclass[10pt,twocolumn,letterpaper]{article}

\usepackage{cvpr}      % To produce the REVIEW version
\usepackage{graphicx}
\usepackage{amsmath}
\usepackage{amssymb}
\usepackage{booktabs}

\usepackage{xcolor} % colors
\usepackage{times}
\usepackage{epsfig}
\usepackage{graphicx}
\usepackage{amsmath}
\usepackage{amssymb}
\usepackage[bb=dsserif]{mathalpha}
\usepackage{bm}

\usepackage{booktabs,colortbl,tabularx}
\usepackage{pifont}%

\usepackage{mathtools}
\usepackage{commath}
\usepackage{algorithm}
\usepackage[noend]{algpseudocode}

\usepackage{multirow}
\usepackage{comment} 
\usepackage{enumitem}

\definecolor{Gray}{gray}{0.9}

\definecolor{battleshipgrey}{rgb}{0.52, 0.52, 0.51}

\usepackage[pagebackref,breaklinks,colorlinks]{hyperref}

\usepackage[capitalize]{cleveref}
\crefname{section}{Sec.}{Secs.}
\Crefname{section}{Section}{Sections}
\Crefname{table}{Table}{Tables}
\crefname{table}{Tab.}{Tabs.}

\def\confName{CVPR}
\def\confYear{2023}

\begin{document}

\title{DiffAVA: Personalized Text-to-Audio Generation with Visual Alignment}

\author{%
  Shentong Mo\\
  Carnegie Mellon University
  \and
  Jing Shi \\
  Adobe Research
  \and
  Yapeng Tian\thanks{Corresponding author.} \\
  University of Texas at Dallas 
}

\maketitle

%%%%%%%%% ABSTRACT
\begin{abstract}

% background
Text-to-audio (TTA) generation is a recent popular problem that aims to synthesize general audio given text descriptions.
% previous methods
Previous methods utilized latent diffusion models to learn audio embedding in a latent space with text embedding as the condition.
% problem
However, they ignored the synchronization between audio and visual content in the video, and tended to generate audio mismatching from video frames.
% solution
In this work, we propose a novel and personalized text-to-sound generation approach with visual alignment based on latent diffusion models, namely DiffAVA, that can simply fine-tune lightweight visual-text alignment modules with frozen modality-specific encoders to update visual-aligned text embeddings as the condition.
Specifically, our DiffAVA leverages a multi-head attention transformer to aggregate temporal information from video features, and a dual multi-modal residual network to fuse temporal visual representations with text embeddings.
Then, a contrastive learning objective is applied to match visual-aligned text embeddings with audio features. 
% results
Experimental results on the AudioCaps dataset demonstrate that the proposed DiffAVA can achieve competitive performance on visual-aligned text-to-audio generation.

\end{abstract}

%%%%%%%%% BODY TEXT

\section{Introduction}

Text-to-Audio (TTA) generation has recently emerged as a popular problem, with the goal of synthesizing high-dimensional audio signals based on given text prompts. Since last year, researchers have explored various denoising diffusion probabilistic models (DDPMs) to learn discrete representations, such as in DiffSound~\cite{yang2022diffsound} and AudioGen~\cite{kreuk2023audiogen}.
Recently, AudioLDM~\cite{liu2023audioldm} proposed training latent diffusion models in a latent space using contrastive language-audio pre-training (CLAP)~\cite{laionclap2023}, with text embeddings serving as the condition. Although these approaches achieve impressive performance in generating plausible sounds, they overlook the synchronization between generated audio and visual content in videos, resulting in misaligned audio and video frames. For instance, the model might generate a train horn sound for ten seconds, even when no train is visible in the frame.

The main challenge is that sounds are naturally aligned with frames in natural videos. 
This inspires us to learn visual-aligned semantics for each text prompt from the video to guide TTA generation. 
To address the problem, our key idea is to capture visual-aligned text representation using lightweight visual-text alignment modules for updating text embeddings as the condition, which differs from existing DDPMs and LDMs on TTA generation. 

We introduce a novel, personalized TTA generation framework called DiffAVA, which is based on LDMs and incorporates visual alignment. DiffAVA generates visual-aligned text embeddings as the condition by simply fine-tuning lightweight visual-text alignment modules with frozen modality-specific encoders. Our framework utilizes a multi-head attention transformer to aggregate temporal information from video representations and a dual multi-modal residual network to fuse temporal visual features with text embeddings. Furthermore, a visual-aligned text-audio contrastive learning objective is employed to match visual-aligned text semantics with audio features. This new framework can support TTA generation that aligns with visual semantics.

Experimental results on the AudioCaps benchmark show that the proposed DiffAVA achieves competitive performance in TTA generation. Moreover, qualitative visualizations with generated audio can demonstrate the success of our DiffAVA framework in visual-aligned TTA generation.

\section{Related Work}

\noindent{\textbf{Diffusion Models.}}
Diffusion models have been demonstrated to be effective in many generative tasks, such as image generation~\cite{saharia2022photorealistic}, image restoration~\cite{saharia2021image}, speech generation~\cite{kong2021diffwave}, and video generation~\cite{ho2022imagen}.
Typically, denoising diffusion probabilistic models (DDPMs)~\cite{ho2020denoising,song2021scorebased} utilized a forward noising process that gradually adds Gaussian noise to images and trained a reverse process that inverts the forward process.
Unlike them, we apply latent diffusion models (LDMs) on audio embeddings to generate personalized sounds aligned with videos based on text descriptions.

\begin{figure*}[t]
    \centering
    % \fbox{\rule{0pt}{2in}
    % \rule{0.8\linewidth}{0pt}}
    \includegraphics[width=0.95\linewidth]{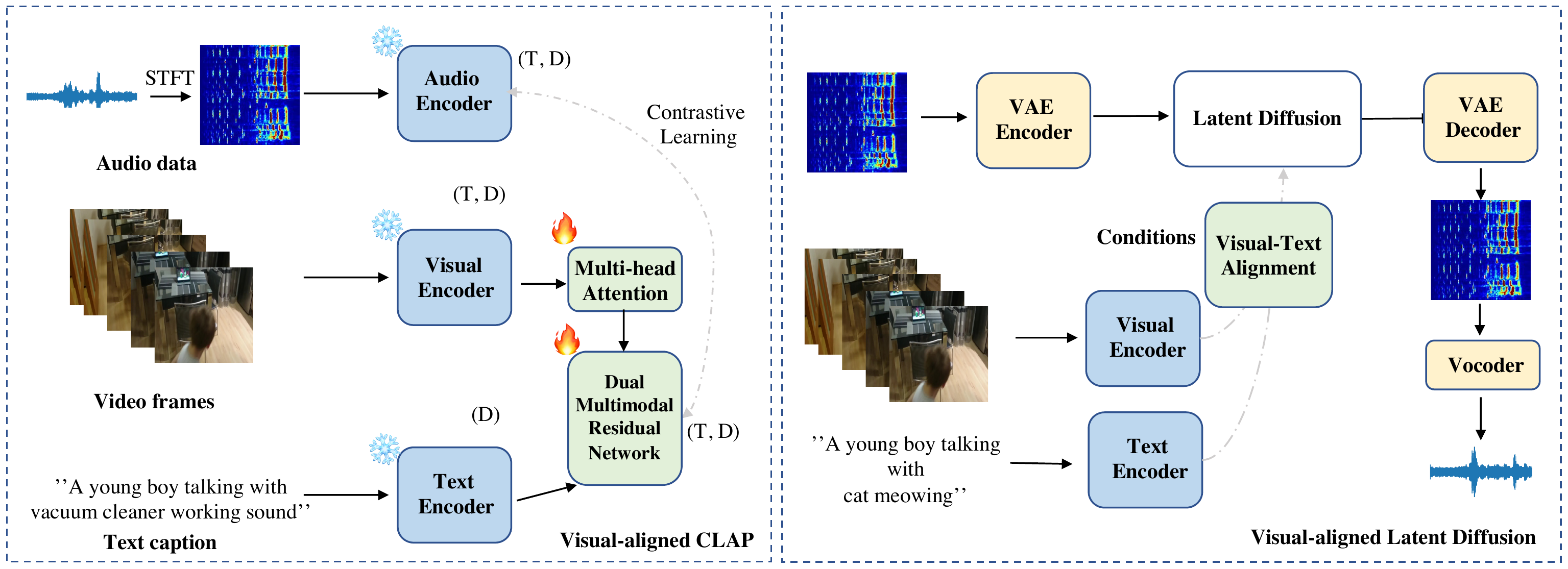}
    \vspace{-0.5em}
    \caption{Illustration of the proposed latent diffusion models for personalized Text-to-audio generation with visual alignment (DiffAVA).
    The multi-head attention transformer aggregates temporal video features, and a dual multi-modal residual network is used to fuse visual representations with text embeddings.
    Then, a contrastive learning objective across each temporal location is applied to match visual-aligned text embeddings with audio features by updating the parameters of our visual-text alignment module. 
    After visual-aligned CLAP pre-training, we directly extract text embeddings with visual-aligned semantics as the condition for latent diffusion models to achieve personalized TTA generation with visual alignment.
    }
    \label{fig: main_img}
    \vspace{-1em}
\end{figure*}

\noindent{\textbf{Audio-Visual Learning.}}
Audio-visual learning has been explored in many previous works to capture the audio-visual alignment between two distinct modalities in videos.
Such cross-modal correspondences are beneficial for many audio-visual tasks, such as audio-event localization~\cite{tian2018ave}, audio-visual parsing~\cite{tian2020avvp,mo2022multimodal}, and audio-visual spatialization \& localization~\cite{Morgado2018selfsupervised,Morgado2020learning,mo2022EZVSL,mo2022SLAVC,mo2023audiovisual,mo2023avsam}.
In this work, our main focus is to learn discriminative cross-modal representations for visual-guided text-to-sound generation, which is more challenging than the abovementioned tasks.

\noindent{\textbf{Text-to-Audio Generation.}}
Text-to-audio (TTA) generation aims to develop the generative model in audio space to  synthesize audio signals based on text prompts.
In recent years, researchers~\cite{yang2022diffsound,kreuk2023audiogen} have tried to explore diverse diffusion models to learn a discrete representation or a discrete space from audio.
More recently, AudioLDM~\cite{liu2023audioldm} applied LDMs to learn the continuous audio representations from latent space in contrastive language-audio pre-training (CLAP), given text embedding as a condition.
However, they failed to generate audio matching from the visual content in the video.
In contrast, we develop a novel and personalized TTA generation approach with visual alignment based on LDMs by efficiently fine-tuning lightweight visual-text alignment modules.

\section{Method}

Given an audio signal from a video and a text prompt, our aim is to synthesize new audio aligned with textual and visual semantics.
We propose a novel TTA generation approach based on LDMs personalized with visual alignment, named DiffAVA, which consists of two main modules, Visual-aligned CLAP in Section~\ref{sec:vclap} and Visual-aligned Latent Diffusion in Section~\ref{sec:vld}.

\subsection{Preliminaries}

In this section, we first describe the problem setup and notations and then revisit conditional latent diffusion models in AudioLDM~\cite{liu2023audioldm} for TTA generation.

\noindent\textbf{Problem Setup and Notations.}
Given audio $a$ and visual frames $v$ from a video and a text prompt $t$, the goal is to generate new audio aligned with textual and visual semantics.
For a video, we have the mel-spectrogram of audio denoted as $\mathbf{A}\in\mathbb{R}^{T\times F}$, and visual frames denoted as $\mathbf{V}\in\mathbb{R}^{T\times H \times W \times 3}$. 
$T$ and $F$ denote the time and frequency, respectively.
For prompts, we have text features $\mathbf{F}^t$ from pre-trained text encoder $f_t(\cdot)$ in CLAP~\cite{laionclap2023}.
For the audio, we use the audio features $\mathbf{F}^a$ from pre-trained text encoder $f_a(\cdot)$ in CLAP~\cite{laionclap2023}.

\noindent\textbf{Revisit AudioLDM.}
To address the TTA generation problem, AudioLDM introduced a conditional latent diffusion model to estimate the noise $\bm{\epsilon}(\bm{z}_n, n, \mathbf{F}^t)$ from the audio prior $\bm{z}_0\in \mathbb{R}^{C\times \frac{T}{r}\times \frac{F}{r}}$ for the mel-spectrogram of audio, where $C$ and $r$ denote the channel of latent representation and the compression level, separately.
For noise estimation, they used the reweighted training objective as
\begin{equation}
    \mathcal{L}_n(\theta) = \mathbb{E}_{\bm{z}_0,\bm{\epsilon},n}\|\bm{\epsilon} - \bm{\epsilon}_\theta(\bm{z}_n, n, \mathbf{F}^a)\|
\end{equation}
where $\bm{\epsilon}\in\mathcal{N}(\bm{0},\bm{I})$ denote the added nosie.
At the final time step $N$ of the forward pass, the input $\bm{z}_n\in\mathcal{N}(\bm{0},\bm{I})$ becomes an isotropic Gaussian noise.
During the training stage, they generated the audio prior $\bm{z}_0$ from the cross-modal representation $\mathbf{F}^a$ of an audio $a$ in a video.
For TTA generation, the text embedding $\mathbf{F}^t$ is used to predict the noise $\bm{\epsilon}_\theta(\bm{z}_n, n, \mathbf{F}^t)$, instead of $\bm{\epsilon}_\theta(\bm{z}_n, n, \mathbf{F}^a)$.

\vspace{-0.5em}
\subsection{Visual-aligned CLAP}\label{sec:vclap}
\vspace{-0.5em}
To align the textual and visual features at spatial and temporal levels corresponding to the paired sound, we apply a multi-head attention transformer to aggregate temporal information from video features $\{\mathbf{F}^v_i\}_{i=1}^T$.
Then, we utilize a dual multi-modal residual network to fuse temporal visual representations with text embeddings $\mathbf{F}^t$ for generating new visual-aligned text embedding $\hat{\mathbf{F}}^t$.
Based on contrastive language-audio pre-training (CLAP), we 
apply visual-aligned CLAP between the textual features with the audio representation in the same mini-batch, which is defined as:
\begin{equation}\label{eq:m2icl}
    \mathcal{L} = 
    - \frac{1}{B}\sum_{b=1}^B \sum_{i=1}^T \log \frac{
    \exp \left( \frac{1}{\tau} \mathtt{sim}(\mathbf{F}^a_{b,i}, \mathbf{F}^v_{b,i}) \right)
    }{
    \sum_{m=1}^B \exp \left(  \frac{1}{\tau} \mathtt{sim}(\mathbf{F}^a_{b,i}, \mathbf{F}^v_{m,i})\right)}
\end{equation}
where the similarity $\mathtt{sim}(\mathbf{F}^a_{b,i}, \mathbf{F}^v_{b,i})$ denotes the temporal audio-textual cosine similarity of $\mathbf{F}^a_{b,i}$ and $\mathbf{F}^v_{b,i}$ across all temporal locations at $i$-th second. 
$B$ is the batch size, $D$ is the dimension size, and $\tau$ is a temperature hyper-parameter.

\vspace{-0.5em}

\subsection{Visual-aligned Latent Diffusion}\label{sec:vld}
\vspace{-0.5em}
With the benefit of visual-aligned CLAP pre-training, we use the pre-trained visual-text alignment module to extract text embeddings $\hat{\mathbf{F}}^t$ with visual-aligned semantics as the condition for latent diffusion models in AudioLDM~\cite{liu2023audioldm}.
Note that we do not fine-tune the parameters of latent diffusion models and directly use the VAE and vocoder released from AudioLDM to achieve efficient and personalized TTA generation with visual alignment.

\vspace{-0.5em}
\section{Experiments}
\vspace{-0.5em}

\subsection{Experimental setup}

\vspace{-0.5em}

\noindent \textbf{Datasets.}
AudioCaps~\cite{kim2019audiocaps} dataset includes 45,423 ten-second audio clips collected from YouTube videos paired with captions for training and 2,240 samples for validation.
Since each audio clip in AudioCaps has 5 text captions, we use the same testing set in AudioLDM~\cite{liu2023audioldm} with 886 instances by selecting one random caption as a text condition.

\noindent \textbf{Evaluation Metrics.}
For comprehensive evaluation between generated audio and target audio, we apply Inception Score (IS), Kullback–Leibler (KL) divergence, Frechet Audio Distance (FAD), and Frechet Distance (FD) as evaluation metrics, following the previous work~\cite{liu2023audioldm}.
IS is used to measure both audio quality and diversity, while KL is evaluated on paired audio.
FAD and FD calculate the similarity between generated audio and reference audio.

\noindent \textbf{Implementation.}
We use the frozen audio and text encoder in AudioLDM~\cite{liu2023audioldm} and only train the parameters of our visual-text fusion module. 
For the video encoder, we apply the pre-trained X-CLIP~\cite{ma2022xclip} as the frozen weights.
The depth of multi-head attention layers with a dimension of 768 is 4, and the number of heads is 8.
The model is trained for 30 epochs using a batch size of 128 and the Adam optimizer with a learning rate of $1.5e-4$.
The latent diffusion model is based on AudioLDM~\cite{liu2023audioldm}, and we use the released weights for VAE and vocoder to generate the final audio samples.

\begin{table}[t]
	%\normalem
	\renewcommand\tabcolsep{6.0pt}
	\centering
	\scalebox{0.85}{
		\begin{tabular}{l|cccc}
			\toprule
			Method & IS ($\uparrow$) & KL ($\downarrow$) & FAD ($\downarrow$) & FD ($\downarrow$)  \\ 	
			\midrule
			DiffSound~\cite{yang2022diffsound} & 4.01 & 2.52 & 7.75 & 47.68 \\
                AudioGen~\cite{kreuk2023audiogen} & -- & 2.09 & 3.13 & -- \\
                AudioLDM~\cite{liu2023audioldm} & 6.90 & 1.97 & \textbf{2.43} & \textbf{29.48} \\		
                DiffAVA (ours) & \textbf{7.37} & \textbf{1.69} & 4.23 & 32.21 \\
			\bottomrule
			\end{tabular}}
   \vspace{-0.5em}
   \caption{Quantitative results of text-to-audio generation on AudioCaps benchmark.}
   \label{tab: exp_sota_audiocaps}
   \vspace{-1.5em}
\end{table}

\vspace{-0.5em}
\subsection{Comparison to prior work}

\vspace{-0.5em}
In this work, we propose a novel and effective framework for text-to-audio generation. 
In order to validate the effectiveness of the proposed DiffAVA, we comprehensively compare it to previous DDPM and LDM baselines:
1) DiffSound~\cite{yang2022diffsound}: a vector-quantized variational autoencoder (VQ-VAE) based DDPM framework by learning a discrete space from audio given natural language description with mask-based text generation.
2) AudioGen~\cite{kreuk2023audiogen} (2023'ICLR):
a recent DDPM approach using a transformer decoder to learn discrete representations from the audio waveform directly. 
3) AudioLDM~\cite{liu2023audioldm} (2023'ICML):
a strong LDM baseline to learn the continuous audio representations from a latent space in contrastive language-audio pre-training.

For TTA generation, we report the quantitative comparison results on the AudioCaps benchmark in Table~\ref{tab: exp_sota_audiocaps}.
As can be seen, we achieve the best results in terms of IS and KL while performing competitively in other metrics.
In particular, the proposed DiffAVA significantly outperforms DiffSound~\cite{yang2022diffsound}, the first DDPM-based baseline on TTA generation, by 3.36 IS, and highly decreases other metrics by 0.83 KL, 3.52 FAD, and 15.47 FD.
Moreover, we achieve decent performance gains of 0.47 IS and a decrease of 0.28 KL compared to AudioLDM~\cite{liu2023audioldm}, the current state-of-the-art TTA generation approach.
These results demonstrate the effectiveness of our approach in learning visual-aligned textual semantics for TTA generation.

\begin{figure}[t]
\centering
% \fbox{\rule{0pt}{2in}
% \rule{0.8\linewidth}{0pt}}
\includegraphics[width=0.99\linewidth]{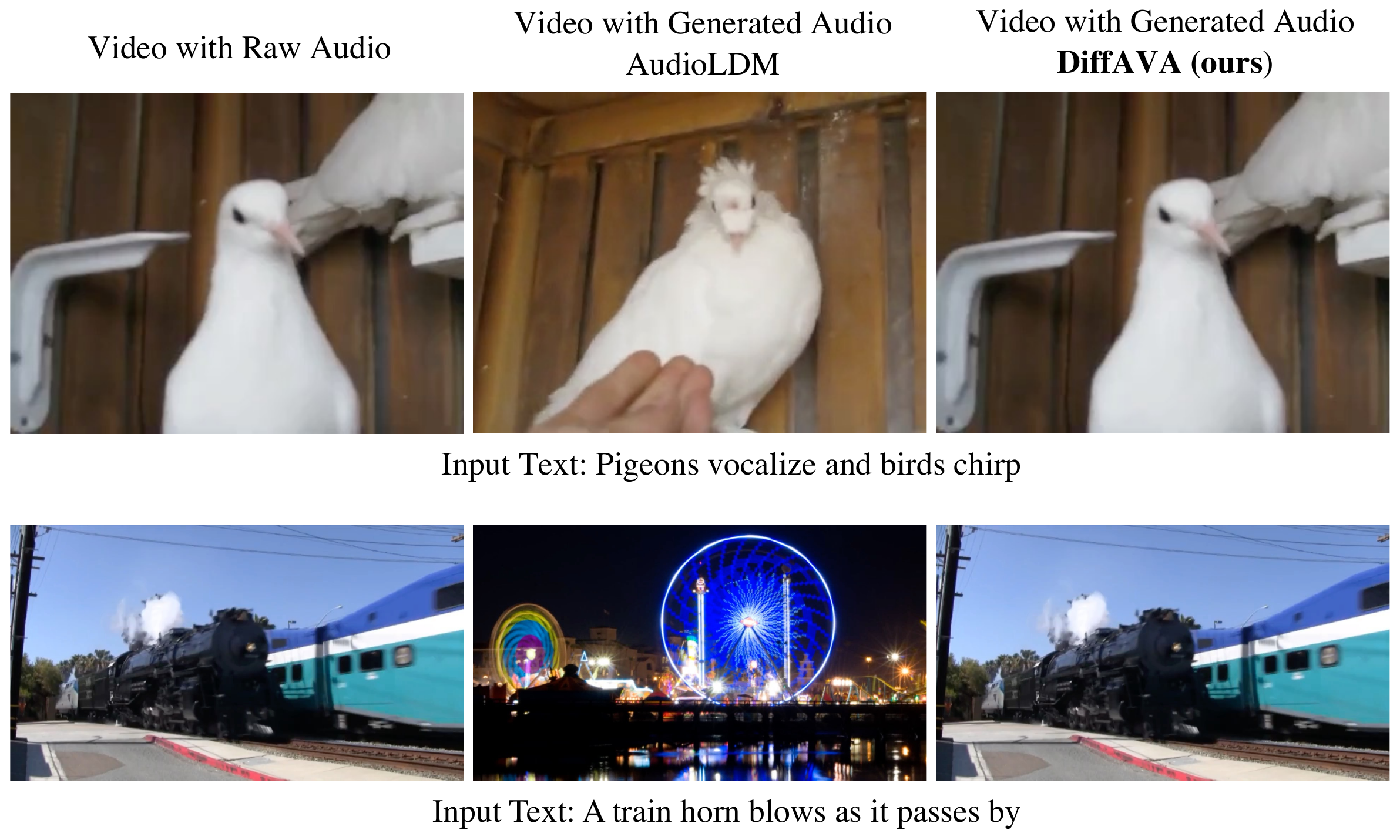}
\vspace{-1em}
\caption{Qualitative comparisons with AudioLDM on visual-aligned TTA generation. 
The proposed DiffAVA produces more accurate and aligned audio for target videos. 
}
\label{fig: vis_cmp}
\vspace{-1.0em}
\end{figure}

In order to qualitatively evaluate the quality of visual-aligned TTA generation, we compare the proposed DiffAVA with AudioLDM~\cite{liu2023audioldm} in Figure~\ref{fig: vis_cmp}.
From comparisons, we observed that without explicit visual-aligned CLAP pre-training objectives, AudioLDM~\cite{liu2023audioldm}, the strong TTA generation baseline fails to discriminate the order of two sounding objects given in the input text, such as ``Pigeons vocalize and birds chirp''.
Meanwhile, it is hard for AudioLDM~\cite{liu2023audioldm} to generate sound temporally aligned with the original video.
For example, given the input text ``A train horn blows as it passes by'', the strong baseline model generates the horn blowing sound across all ten seconds although no train appeared in the last several seconds. 
In contrast, the generated audio from our method is more aligned with the visual semantics existing in the video.
These visualizations further showcase the superiority of our simple DiffAVA in visual-aligned TTA generation.

\vspace{-0.5em}
\section{Conclusion}
\vspace{-0.5em}
In this work, we present DiffAVA, a novel and personalized text-to-sound generation approach with visual alignment based on latent diffusion models, that can simply fine-tune lightweight visual-text alignment modules with frozen modality-specific encoders to generate text embeddings with visual-aligned semantics as the condition.
We introduce a multi-head attention transformer in CLAP to aggregate temporal information from video features and a dual multi-modal residual network to fuse temporal visual embeddings with text features.
Then, we leverage contrastive learning to learn the correspondence between audio features and text embeddings with visual-aligned semantics. 
% results
Empirical experiments on the AudioCaps dataset demonstrate the state-of-the-art performance of our DiffAVA on visual-aligned text-to-audio generation.

%%%%%%%%% REFERENCES
{\small
\bibliographystyle{ieee_fullname}
\bibliography{reference}
}

\end{document}